\documentclass{article} 
\usepackage[preprint]{colm2026_conference}
\usepackage{microtype}
\usepackage{hyperref}
\usepackage{url}
\usepackage{booktabs}
\usepackage{multirow}
\usepackage{graphicx}
\usepackage{wrapfig}
\usepackage[table,xcdraw,usenames,dvipsnames]{xcolor}

\usepackage{lineno}
\usepackage{enumitem}
\setlist[itemize]{leftmargin=*, noitemsep, topsep=0pt}
\setlist[enumerate]{leftmargin=*, noitemsep, topsep=0pt}

\definecolor{darkblue}{rgb}{0, 0, 0.5}
\definecolor{rowshade}{HTML}{F0F4FA}
\definecolor{bestcell}{HTML}{D6EAF8}
\definecolor{hdrblue}{HTML}{2C3E6B}
\definecolor{secgray}{HTML}{5D6D7E}

\newcounter{cfinding}
\newenvironment{cfinding}[1][]{\refstepcounter{cfinding}\par\medskip
   \noindent\textsc{Takeaway~\thecfinding. #1} \rmfamily}{\medskip}

\usepackage[most]{tcolorbox}

\usepackage{tabularx}
\usepackage{amsfonts}
\usepackage{cleveref}

\newtcolorbox{PromptBox}[1]{
    enhanced, breakable, colback=gray!2, colframe=black!80,
    fonttitle=\bfseries\sffamily, title=#1, sharp corners,
    boxrule=0.7pt, attach boxed title to top left={yshift=-2mm, xshift=3mm},
    boxed title style={colback=black!80, sharp corners}
}

\newtcolorbox{CodeBox}[1]{
    enhanced, colback=black!90, colframe=black, coltitle=white,
    fonttitle=\small\ttfamily, title=#1, sharp corners,
    boxrule=0.5pt, fontupper=\small\color{white}\ttfamily
}

\hypersetup{colorlinks=true, citecolor=darkblue, linkcolor=darkblue, urlcolor=darkblue}

\title{Decomposing the Delta: What Do Models Actually Learn from Preference Pairs?}

\author{
  \textbf{Chia-Hsuan Lee, Mingyang Zhou, Renkun Ni, Zelei Cheng, Sihui Dai,} \\
  \textbf{Supriyo Chakraborty, Shixiong Zhang, Sambit Sahu, William Campbell} \\
  Capital One
}
\begin{document}
\tcbset{
    takeawayboxstyle/.style={
        colback=gray!10, 
        colframe=white, 
        fonttitle=\bfseries, 
        coltitle=black, 
        left=1mm, 
        right=1mm,
        top=0.5mm,
        bottom=0.5mm,
        enhanced, 
        borderline west={0.2mm}{0pt}{Green}, 
        boxrule=0mm, 
        sharp corners, 
        before skip=5pt,
        after skip=5pt, 
    }
}

\ifcolmsubmission
\linenumbers
\fi

\maketitle

\begin{abstract}

 Preference optimization methods such as DPO and KTO are widely used for aligning language models, yet little is understood about what properties of preference data drive downstream reasoning gains.  We ask: \textit{what aspects of a preference pair improve a reasoning model's performance on general reasoning tasks?} We investigate two distinct notions of quality delta in preference data: \textit{generator-level delta}, arising from the differences in capability between models that generate chosen and rejected reasoning traces, and \textit{sample-level delta}, arising from differences in judged quality differences within an individual preference pair. To study generator-level delta, we vary the generator's scale and model family, and to study sample-level delta, we employ an LLM-as-a-judge to rate the quality of generated traces along multiple reasoning-quality dimensions.  We find that increasing generator-level delta steadily improves performance on out-of-domain reasoning tasks and filtering data by sample-level delta can enable more data-efficient training.  Our results suggest a twofold recipe for improving reasoning performance through preference optimization: maximize generator-level delta when constructing preference pairs and exploit sample-level delta to select the most informative training examples.
\end{abstract}

\section{Introduction}

Preference optimization methods such as DPO \citep{rafailov2023direct} and KTO \citep{ethayarajh2024kto} have become central to aligning language models with human preferences. In standard alignment pipelines, a model learns from paired responses---one chosen, one rejected---ranked by human annotators or LLM judges \citep{ouyang2022training, bai2022training}. When applied to reasoning tasks, these pairs are typically constructed by contrasting a solution that reaches the correct final answer against one that does not, treating outcome correctness as the primary signal for learning \citep{lai2024stepdpo,lightman2023lets}.

Recently, the delta learning hypothesis \citep{geng2025delta} challenged the assumption that high-quality supervision is necessary for effective preference tuning. \citet{geng2025delta} showed that the relative quality gap---the \emph{delta}---between two weak model responses suffices to improve a stronger student, even when neither response exceeds the student's own capability. This result reframes the central question in preference data construction: what matters is not the absolute quality of either response, but the gap between them. 

While \citet{geng2025delta} analyzed delta learning on open-ended response tasks on instruction-tuned models, we focus on analyzing delta learning in reasoning models and reasoning tasks.  In this setting, the ``delta" in quality between chosen and rejected responses is less well captured by common notions of preference such as ``helpfulness" and ``instruction-following" \citep{lambert2024tulu3}.  In this work, we analyze how different notions of quality across chosen and rejected reasoning traces can impact performance in downstream verifiable domains---mathematical reasoning, STEM, and code.


We anchor our analysis around two notions of delta. The first, which we call \textbf{generator-level delta}, is inherited from prior work: it is the capability gap between the models that produce chosen and rejected responses---a design choice made during data construction. The second, which we introduce, is \textbf{sample-level delta}: the quality difference within a single preference pair, measured along specific reasoning dimensions such as factuality, coherence, and computational precision. Prior work operates primarily at the generator level and does not perform a fine-grained analysis of different characterizations of sample-level delta. Our goal is to characterize the sample-level signal that generator-level choices produce and to understand which dimensions of that signal drive downstream reasoning gains.

Our investigation proceeds in three parts.

\textbf{First, we verify that delta learning transfers to verifiable reasoning and study its scaling behavior (\S\ref{sec:delta_reasoning})}. Using preference pairs from models spanning two families (Nemotron-4B--49B and s1-3B--32B), we confirm that weak-model deltas remain effective: training on pairs from s1-7B vs.\ s1-3B---without any ground-truth verification---matches DPO trained on DeepSeek-R1 data with outcome verification. We then scale the generator-level delta by increasing the capability of the chosen model. Perhaps surprisingly, in-domain math performance is largely invariant to this scaling, while out-of-domain performance on STEM and code improves steadily. We trace this divergence to sample-level deltas: as the chosen model scales, the quality gap between chosen and rejected responses widens across all five reasoning dimensions we measure, and it is this widening that explains the out-of-domain gains.

\textbf{Second, we ask what the sample-level delta actually encodes (\S\ref{sec:what_delta_encodes})}. The most natural hypothesis is that it is driven by outcome correctness: the chosen response reaches the right answer and the rejected one does not. We test this directly by constructing preference pairs across all four correctness combinations---correct-vs-incorrect, correct-vs-correct, incorrect-vs-incorrect, and incorrect-vs-correct---and training with DPO on each. Surprisingly, in-domain performance is largely invariant to correctness labels. Even incorrect-vs-correct pairs yield gains over the base model. This rules out outcome correctness as the dominant signal and points to subtler properties of the reasoning process itself.

\textbf{Third, we decompose the sample-level delta into interpretable quality dimensions (\S\ref{sec:quality_decomposition})}. Motivated by the insufficiency of outcome correctness, we evaluate each chain-of-thought trace along five axes---factuality, strategy coherence, step coherence, computational precision, and signal-to-noise ratio---using a calibrated LLM judge. By selecting subsets of pairs with the largest sample-level deltas in each dimension, we isolate which components of reasoning quality the model learns from. We find that step coherence is the most consistently effective selection criterion, and that training on as few as 5k high-delta pairs matches or exceeds performance from the full 16.5k set---demonstrating that sample-level data selection can substitute for brute-force scaling of data or generators.

\section{Decomposing the Delta}
\label{sec:decomposing_delta}
The delta learning hypothesis~\citep{geng2025delta} establishes that a quality gap between paired responses is sufficient for preference optimization to improve a student model. But ``quality gap'' is not a single quantity---it can be defined at different levels of granularity, and the level at which it is defined determines what questions can be asked about it. In this section, we formalize two notions of delta that structure the remainder of our investigation, and describe how we measure the finer-grained of the two.
 
\subsection{Two Notions of Delta}
\label{sec:delta_definitions}
 
\textbf{Generator-level delta.}
The most direct way to create a quality gap is to pair outputs from models of different capability. We define the \emph{generator-level delta} as the capability difference between the model that produces chosen responses and the model that produces rejected responses. Prior work on delta learning~\citep{geng2025delta} operates entirely at this level, varying the generator pairing and measuring downstream gains as a function of the gap.  We will study this notion of delta for reasoning tasks in Section \ref{sec:scaling_generator}.
 
\textbf{Sample-level delta.}
Generator-level delta tells us how preference data was constructed, but not what any particular preference pair contains. Two pairs drawn from the same generator pairing can differ enormously. We define the \emph{sample-level delta} as the quality difference within an individual chosen--rejected pair, measured along specific reasoning dimensions.
\subsection{Measuring Sample-Level Delta}
\label{sec:reasoning_quality_dimensions}
In this paper, we consider several different methods for measuring differences in sample level delta which we outline here.

\textbf{Outcome correctness. }A simple measure of sample-level delta is measuring differences in the correctness of the final answer from the generated trace.  In Section \ref{sec:outcome_correctness}, we will investigate the informativeness of this measure of sample-level delta by deciding chosen and rejected samples based on correctness.

\textbf{Reasoning trace quality. }A more fine-grained notion of sample-level delta for reasoning tasks is differences in quality of the reasoning itself rather than just the correctness of the final answer.  In order to assess reasoning trace quality, we utilize LLM-as-a-judge to rate generated reasoning traces across several dimensions following \citet{lee2025evaluating}:
\begin{itemize}
    \item \textbf{Factuality.} Factuality is the grounding of each trace to the given problem and specifically checks whether the response properly uses values and constraints provided in the problem instead of hallucinating.
    \item \textbf{Coherence.} We consider 2 measures of coherence, which we will refer to as \textit{Strategy Coherence} and \textit{Step Coherence}. Strategy Coherence measures whether or not the overall plan for solving the problem is reasonable for addressing the problem, while Step Coherence investigates whether steps in the reasoning logically flow from the previous.
    \item \textbf{Computational Precision.} This dimension measures the degree to which the trace makes numerical computation errors such as arithmetic errors and sign errors during the reasoning process.
    \item \textbf{Signal-to-Noise.} This dimension measures to what degree the trace contains relevant calculations and reasoning steps (signal) instead of unnecessary redundant calculations, irrelevant digressions, and verbose meta-commentary.
\end{itemize}
LLM-as-a-judge ratings along each dimension allow us to have a quantitative measure of sample-level delta across chosen an rejected pairs.  In Section \ref{sec:scaling_generator}, we will investigate the connection between generator-level delta and sample-level delta with respect to each of these dimensions.  Additionally, in Section \ref{sec:quality_decomposition}, we will investigate the importance of sample-level deltas across each dimension by using delta magnitudes as a means for selecting high quality preference pairs.

\section{Experimental Setup}
\subsection{Dataset and Evaluation}

\subsubsection{Training Data}

We build our training data from OpenR1-Math-220k~\citep{openr1}, a large-scale mathematical reasoning dataset comprising 220k problems sourced from NuminaMath~1.5~\citep{numina2024}, each accompanied by two to four reasoning traces generated by DeepSeek-R1~\citep{deepseekr1}. Trace correctness is verified via Math-Verify for the majority of samples and Llama-3.3-70B-Instruct \citep{meta-llama-3.3-70b-instruct} as a judge for the remainder. We use the \texttt{default} split (94k problems), which has been reported to yield the strongest performance after supervised finetuning. To construct preference pairs with controlled correctness properties, we first filter problems that contain at least one verified-correct response and at least one verified-incorrect response among their generated traces. From this filtered pool, we randomly sample 16.5k problems to form our training set. More training details can be found in in Appendix \ref{app:training_details}.

\subsubsection{Evaluation}

We evaluate across three domains: mathematical reasoning, advanced STEM knowledge, and code generation, where mathematical reasoning is used for in-domain evaluation and the remaining for out-of-domain evaluation.

\textbf{Mathematical reasoning. } We partition math benchmarks into two difficulty tiers. The \textbf{Easy} group comprises MATH-500~\citep{hendrycksmath2021,lightman2023lets}, GSM8K~\citep{cobbe2021gsm8k}, and AMC~2023~\citep{maa2023amc}. The \textbf{Hard} group comprises Minerva-Math~\citep{lewkowycz2022minerva}, OlympiadBench~\citep{he2024olympiadbench}, AIME~2024~\citep{aime24}, and AIME~2025~\citep{aime25}. 

\textbf{Out-of-domain evaluation. } To assess whether delta learning transfers beyond the mathematical reasoning training distribution, we evaluate on MMLU-Pro~\citep{wang2024mmlupro} for expert-level multi-disciplinary knowledge and reasoning, evaluate on TheoremQA~\citep{chen2023theoremqa} for theorem questions spanning mathematics, physics, and engineering. We additionally report results on LiveCodeBench~\citep{jain2024livecodebench} for code generation.

\textbf{Inference. } All inference runs use temperature $0.7$ and top-$p = 1.0$. For AMC~2023, AIME~2024, and AIME~2025---where the evaluation sets are small and single-sample variance is high---we report avg@16 (the average accuracy across 16 independent samples per problem). All other benchmarks are reported as pass@1.

\textbf{Rating reasoning quality per sample. } We prompt GPT-OSS-120b~\citep{openai2025gptoss120bgptoss20bmodel} to provide a rating between 1-5 for each criteria listed in Section \ref{sec:reasoning_quality_dimensions} with medium reasoning effort and average across 5 generations at temperature 0.6. We provide additional details about our judge setup in Appendix \ref{app:trace_judge_setup}.


\subsection{Models}
\subsubsection{Base Model}
For all experiments, we use Nemotron-8B~\citep{bercovich2025llama} as the base policy model to verify different ``delta" for reasoning preference tuning. 

\subsubsection{Response Synthesis Models}
\label{subsection:response_synth}
We generate reasoning traces and final answers from multiple model families spanning a range of parameter scales. In particular, we consider Nemotron~\citep{bercovich2025llama} models with sizes from 4B to 49B parameters and s1~\citep{muennighoff2025s1} models with sizes from 3B to 32B parameters. For each candidate model, we use prompts sampled from the OpenR1 dataset~\citep{openr1} and collect eight reasoning traces per prompt, keeping the generation configuration fixed within each model family. Following the OpenR1 protocol, we prepend each user prompt with the instruction: ``Please reason step by step, and put your final answer within \texttt{\textbackslash boxed\{\}}.''. These synthesis models provide a diverse pool of candidate reasoning trajectories that are later used to construct verified responses and preference pairs. Comprehensive performance results for these models can be found in Appendix~\ref{app:benchmark-comparison}.
\vspace{-10pt}
\section{Does Delta Learning Work for Reasoning?}
\label{sec:delta_reasoning}

\subsection{Delta Learning vs. Outcome Verified Preference}

\begin{wrapfigure}{r}{0.55\textwidth}
 \centering
 \includegraphics[height=6cm, keepaspectratio]{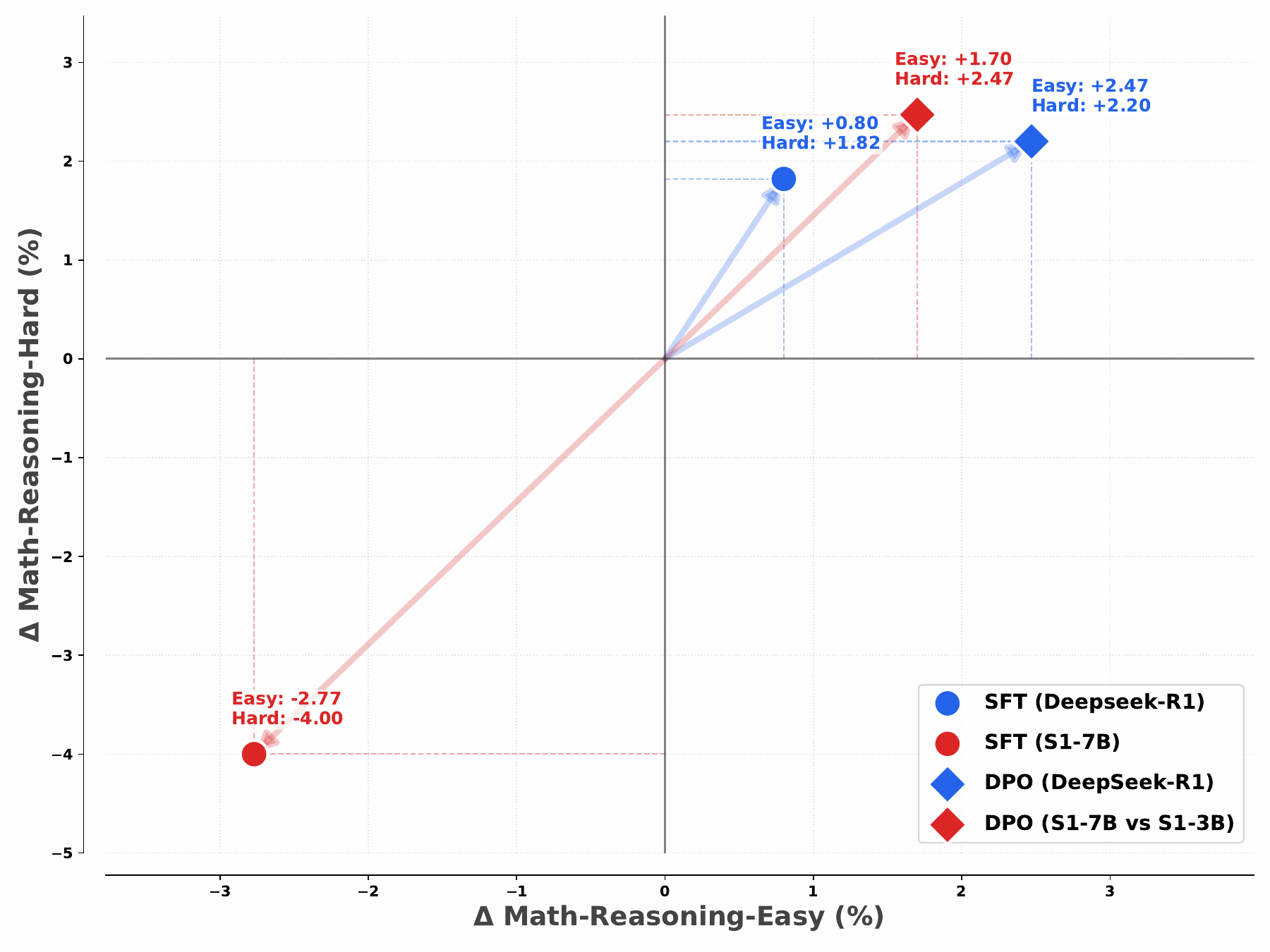}
  \caption{The relevant performance gain against base model (Nemotron-8B) over math-reasoning easy and math-Reasoning hard evaluation sets after SFT and DPO fine-tuning on responses from the weak model (S1) and the strong model (Deepseek-R1)}
  \label{fig:delta_vs_outcome}
\end{wrapfigure}

We extend the study of the \textit{Delta Learning Hypothesis} to reasoning models and tasks. Compared with standard instruction-following settings, reasoning tasks require models to produce coherent multi-step inference traces, making it less clear whether weak preference pairs can still provide effective supervision. To evaluate the hypothesis in this more challenging setting, we examine whether preference optimization on reasoning data generated by weaker models can nevertheless improve reasoning performance. For the strong-data setting, we construct preference pairs from OpenR1-Math, where all responses are generated by DeepSeek-R1. We use an outcome verifier to assess the correctness of each response, treating correct generations as chosen and incorrect generations as rejected. For the weak-data setting, we use the \texttt{s1-7B} and \texttt{s1-3B} generations described in Section~\ref{subsection:response_synth}, regarding \texttt{s1-7B} responses as chosen and \texttt{s1-3B} responses as rejected. In addition, to highlight the effect of preference learning, we perform SFT on the chosen responses from both paired preference datasets. As shown in Figure~\ref{fig:delta_vs_outcome}, one epoch of SFT on \texttt{s1-7B}-generated data degrades model performance, while one epoch of SFT on DeepSeek-R1-generated data improves it. In contrast, although the two generators differ substantially in reasoning ability, DPO training on their corresponding preference pairs leads to similar downstream reasoning performance, and both DPO-trained models outperform the original baseline. These findings provide evidence that the \textit{Delta Learning Hypothesis} also holds in reasoning tasks.

\subsection{Scaling Law of Delta Learning}
\label{sec:scaling_generator}
To investigate the scaling properties of delta learning, we establish a controlled setup by fixing the model used to generate rejected responses (s1-3B) while scaling the capability of the model used to generate chosen responses. We perform training with DPO and KTO respectively. We evaluate this scaling effect across two primary axes: \textit{downstream benchmark performance} and \textit{quality of the generated preference data}.

\begin{figure}[]
 \centering
  \includegraphics[width=\linewidth]{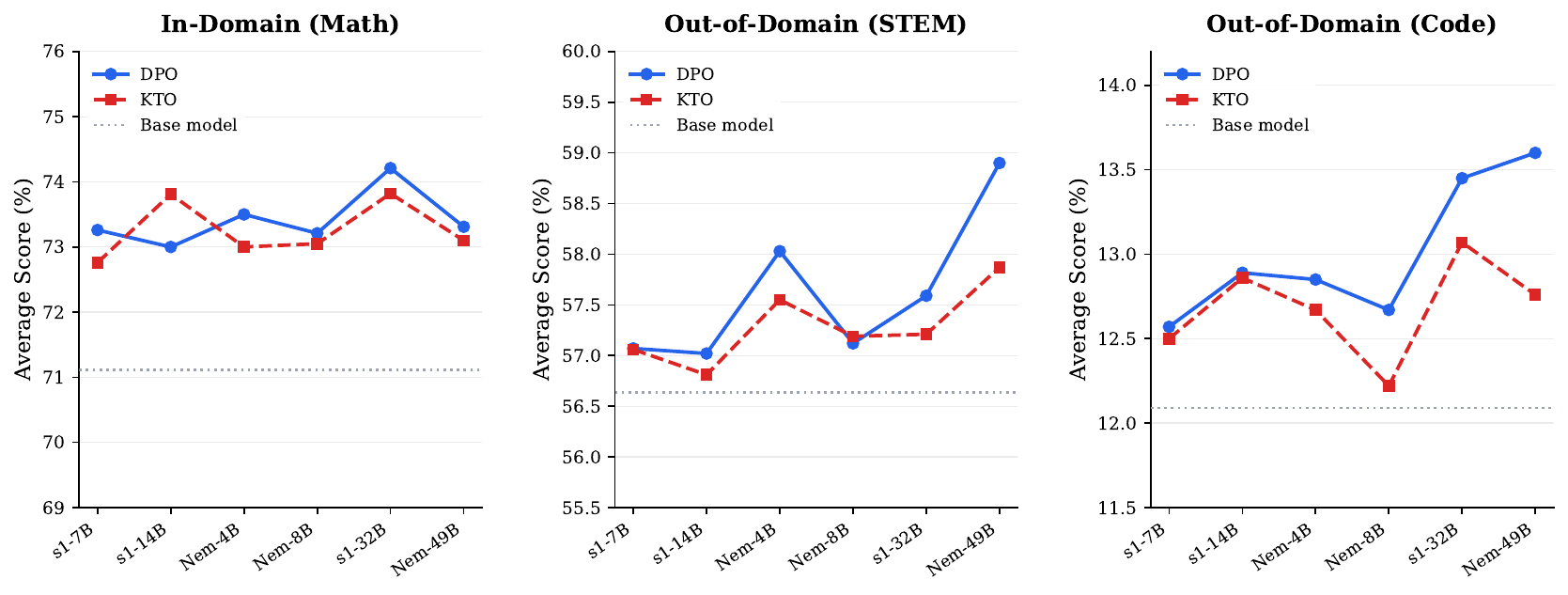}
  \caption{Scaling law of delta learning for reasoning tasks. Chosen models are ordered by off-the-shelf math capability. All models use s1-3B as the rejected response.}
  \label{fig:scalinglaw}
\end{figure}

\textbf{Downstream benchmark performance. }The evaluation results of the impact for downstream performance on general reasoning benchmarks are presented in Figure \ref{fig:scalinglaw}. 
Interestingly, because our training utilizes the math-centric OpenR1 dataset, we observe only a weak scaling effect on in-domain mathematical benchmarks. We hypothesize that this plateau is a result of the base model's pre-existing proficiency. When a strong base model is already highly capable in a specific domain, extracting further significant performance gains becomes inherently difficult.

In contrast, delta learning yields substantial improvements on out-of-domain tasks, particularly in STEM and coding evaluations. Performance on broader reasoning benchmarks, such as LiveCodeBench, increases noticeably. Crucially, the magnitude of this improvement correlates directly with the generator-level delta. A larger disparity between the two models consistently translates into stronger out-of-domain generalization.\footnote{Given that DPO consistently outperformed KTO in preliminary evaluations, we employ DPO for all subsequent experiments.}

\textbf{Quality of the generated preference data. }To understand the mechanics behind these out-of-domain improvements, we analyze how generator-level delta relates to sample-level delta measured by the reasoning trace quality dimensions defined Section \ref{sec:reasoning_quality_dimensions} for the generated preference pairs. In Table~\ref{tab:scaling_quality}, we report the average across the dataset of sample-level deltas measured in each dimension.

The results reveal a clear trend: as the capability of the chosen model scales (e.g., from s1-7B to s1-32B, or Nemotron-4B to Nemotron-49B), the sample-level delta across all measured dimensions widens significantly. For instance, the advantage in step coherence jumps from 0.96 (s1-7B vs. s1-3B) to over 2.0 (s1-32B and Nemotron-49B vs. s1-3B). Similar scaling trajectories are observed for factuality and strategy coherence.

This analysis confirms that scaling the chosen model does not merely change the generated outputs; it systematically enriches the quality and density of the training signal. This enriched signal-to-noise ratio directly explains the enhanced out-of-domain generalization observed in our benchmark evaluations, proving that a larger capability gap in the preference pairs is instrumental for teaching broader reasoning skills.

\begin{tcolorbox}[takeawayboxstyle]
    \begin{cfinding}
    Scaling the capability gap between chosen and rejected generators yields diminishing returns in-domain but steadily improves out-of-domain generalization on STEM and code, with the quality delta across all reasoning dimensions widening as the chosen model scales.
    \end{cfinding}
\end{tcolorbox}

\begin{table}[]
    \small
    \centering
    \renewcommand{\arraystretch}{1.25}
    \setlength{\tabcolsep}{5pt}
    \resizebox{\columnwidth}{!}{%
    \begin{tabular}{
        >{\raggedright\arraybackslash}p{3.6cm}
        c c c c c}
    \toprule
    \rowcolor{hdrblue}
    \textcolor{white}{\textbf{Chosen $\backslash$ Rejected}}			
      & \textcolor{white}{\textbf{step coherence}}
      & \textcolor{white}{\textbf{factuality}}
      & \textcolor{white}{\textbf{computational precision}} 
      & \textcolor{white}{\textbf{strategy coherence}}
      & \textcolor{white}{\textbf{signal-to-noise}}\\
    \midrule
    \rowcolor{hdrblue!15}
    s1-7B vs s1-3B              & 0.96	&	1.06	&	0.85	&	0.92	&	0.53 \\
    \rowcolor{rowshade}
    s1-14B vs s1-3B             & 1.6	&	1.54	&	1.39	&	1.44	&	0.94 \\
    Nemotron-4B vs s1-3B        & 1.77	&	1.68	&	1.56	&	1.54	&	0.74 \\
    \rowcolor{rowshade}
    Nemotron-8B vs s1-3B        & 2.01	&	1.7	 &	1.71	&	1.74	&	1.18\\
    s1-32B vs s1-3B             & 2	& 1.71	& 1.67 & 1.76	&	1.21 \\
    \rowcolor{rowshade}
    Nemotron-49B vs s1-3B       & 2.08	&	1.77	&	1.69	&	1.81	&	1.34\\
    \bottomrule
    \end{tabular}
    }
    \caption{Quality dimension score difference for the scaling laws experiments on general reasoning benchmarks. All models use s1-3B as the rejected response.}
    \label{tab:scaling_quality}
\end{table}
\section{What Does Delta Actually Encode?}
\label{sec:what_delta_encodes}
\subsection{Outcome Correctness}
\label{sec:outcome_correctness}
A natural assumption in reasoning preference optimization is that the sample-level delta
between paired responses is primarily driven by outcome correctness: the chosen
response reaches the right answer, and the rejected response does not. We test
this assumption by constructing preference pairs across all four correctness
combinations---correct-vs-incorrect, correct-vs-correct, incorrect-vs-incorrect,
and incorrect-vs-correct---using responses sampled from the same model
(Nemotron-4B) on mathematical reasoning tasks, and training Nemotron-8B with DPO
on each configuration. To ensure a controlled comparison, all four correctness configurations are derived from the same underlying set of 3.5k problems.

\begin{figure}[t]
 \centering
\includegraphics[height=7cm, keepaspectratio]{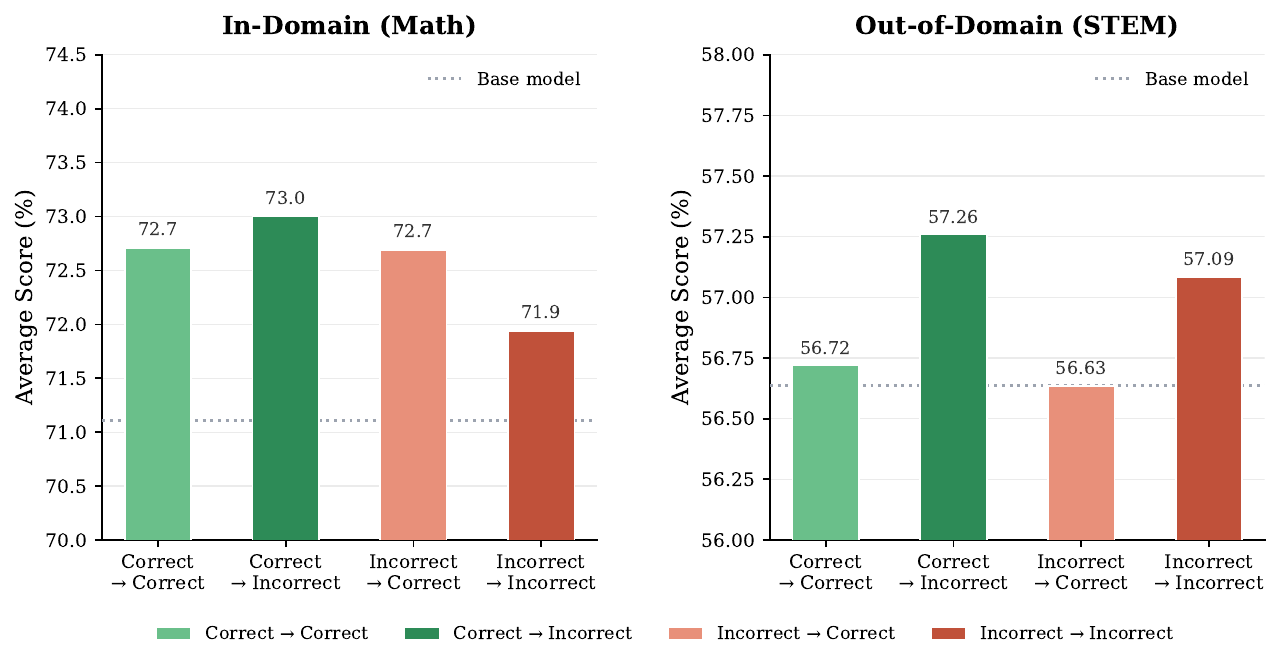}
\caption{Nemotron-8B DPO performance on all correctness combinations (chosen → rejected). All pairs generated by Nemotron-4B.}
  \label{fig:correctness}
\end{figure}

We present our results in Figure~\ref{fig:correctness}. On in-domain
mathematical reasoning, all four correctness combinations improve over the base
model ($\sim$71.1\%), with scores ranging from 71.9\% to 73.0\%. The
conventional correct-vs-incorrect configuration performs best, and
incorrect-vs-incorrect performs worst---so correctness labels are not
irrelevant. However, the gap between the best and worst configurations is modest
(1.1 points), and notably, even incorrect-vs-incorrect pairs still yield gains
over the base model. This suggests that outcome correctness is \emph{one}
contributing factor to the delta, but not the dominant one.

On out-of-domain STEM tasks, we observe a similar overall pattern---all
configurations improve over or match the base model---though a slight trend
emerges. Configurations where the rejected response is incorrect
(correct$\to$incorrect: 57.26\%; incorrect$\to$incorrect: 57.09\%) tend to
modestly outperform those where the rejected response is correct
(correct$\to$correct: 56.72\%; incorrect$\to$correct: 56.63\%). This pattern
is consistent with the intuition that learning to avoid flawed reasoning may
be somewhat more transferable than learning to imitate correct solutions,
echoing findings from~\citep{razin2025unintentional,cho-etal-2025-rethinking} that DPO's gradient
dynamics are often dominated by the ``pushing away'' from rejected responses
rather than the ``pulling toward'' chosen ones.



\begin{tcolorbox}[takeawayboxstyle]
  \begin{cfinding}
    Outcome correctness contributes to but does not dominate delta learning for
    reasoning. All four correctness combinations---including
    incorrect-vs-correct---yield gains over the base model on in-domain math. This indicates that a significant portion of the learning signal resides in reasoning process quality rather than answer
    correctness alone.
  \end{cfinding}
\end{tcolorbox}


\subsection{Reasoning Chain-of-Thought Quality and Sample Efficiency} \label{sec:quality_decomposition}
\begin{figure}[th]
 \centering
  \includegraphics[width=\linewidth]{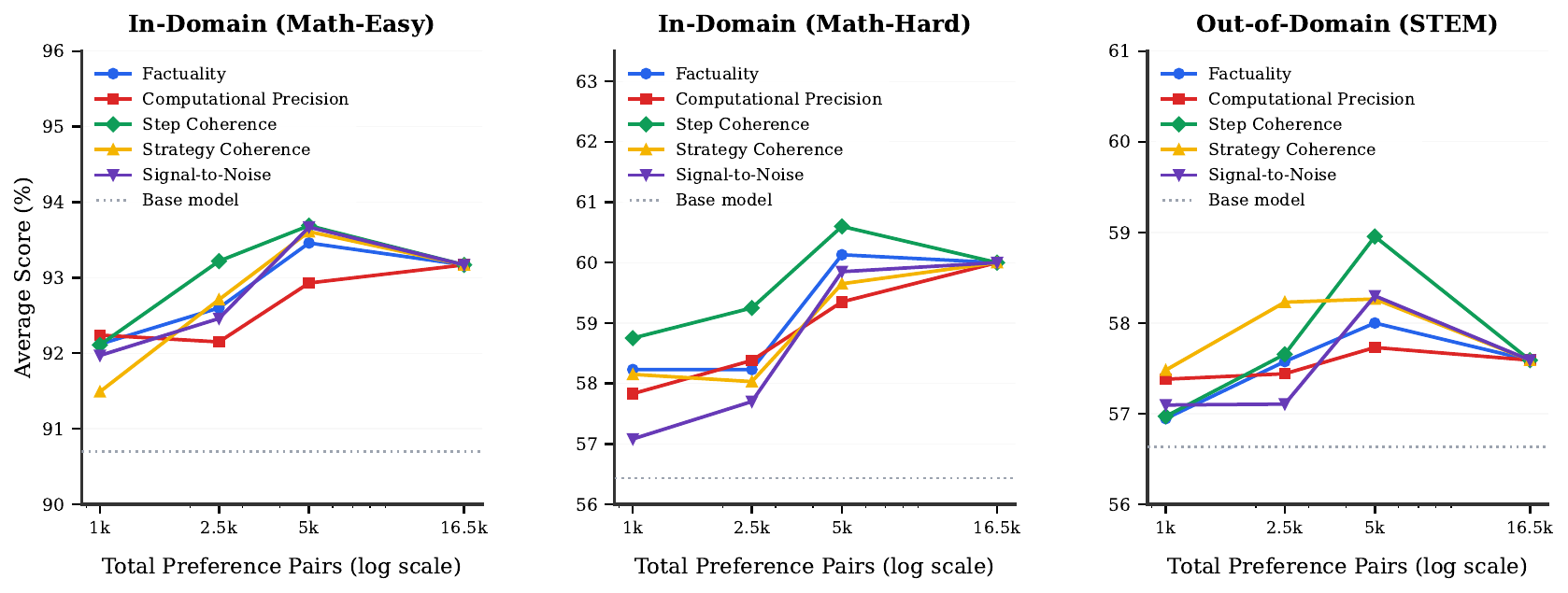}
  \caption{\textbf{Scaling performance across quality dimensions for S1-32B vs. S1-3B.} We show accuracy as a function of the top-k pairs with highest quality difference measured by each quality dimension (with 16.5k representing the full set of preference pairs). Performance is evaluated on in-domain (easy and hard math) and out-of-domain (STEM) tasks.  The dotted grey line represents the base model performance without DPO.} 
  \label{fig:dimension_scale}
\end{figure}

If outcome correctness does not explain the effectiveness of delta learning for reasoning, then the operative signal must reside in the reasoning process itself. We hypothesize that the most informative sample-level deltas reflect intrinsic differences in chain-of-thought quality — differences that persist even when two responses share the same final answer. In this section, we conduct ablation studies using delta preference pairs from S1-32B vs. S1-3B. We take top 1k, 2.5k, and 5k pairs that have the largest delta in each quality dimension and train with DPO. We provide an analysis of overlaps between these sets in Appendix \ref{app:dim_overlap}.  We present our results in Figure \ref{fig:dimension_scale}.

\textbf{Early gains from limited data. }From Figure \ref{fig:dimension_scale}, we observe that DPO can consistently improve performance over the base model on both in-domain and out-of-domain tasks with as few as 1k selected preference pairs.  These gains are especially pronounced for in-domain tasks.  For example, using the top 1k pairs yields improvements in accuracy of up to 2\% on easy math tasks and 2.32\% on hard math tasks.  On STEM tasks, we find that the gains are smaller (up to 0.845\% when top-1k in strategy coherence are chosen), but improvement is consistent across all 1k sample sets.

Additionally, we find that as the number of selected pairs increases, the marginal benefit of adding more data decreases. Across both in-domain and out-of-domain tasks, we find that DPO with top-5k preference pairs already achieves performance comparable to and can, in some settings, outperform DPO with the full 16.5k preference set. For example, on the task of hard math problems, training with the full preference set achieves 60\% accuracy while 5k subsets achieve within $\pm$0.65\% of this value.  This suggests that the full preference set contains many pairs that are less informative than the selected top-k subset of highest-delta examples across each dimension, and much of the benefit of preference optimization can be recovered from a subset of carefully selected data.

\textbf{Step coherence as a data selection criterion. }We observe that selecting pairs based on the magnitude of their reasoning-quality delta can improve data efficiency in DPO. Among reasoning quality dimensions considered, \textit{step coherence} appears to be the most consistently effective selection criterion: across subset size and evaluation domains, selecting pairs with the largest deltas in step coherence generally yields the best or near-best downstream performance. These results suggest that differences in step coherence capture an important component of the preference signal exploited by DPO, making it a strong heuristic for selecting high-value preference pairs.

\begin{tcolorbox}[takeawayboxstyle]
    \begin{cfinding}
    Much of the benefit of DPO for reasoning tasks can be recovered by using a small high quality subset of preference pairs chosen to have large reasoning-quality deltas. Among all reasoning dimensions, we find that step coherence is the most consistently effective measure of reasoning-quality for data selection.
    \end{cfinding}
\end{tcolorbox}


\section{Related Work}
\textbf{Learning from preference feedback. }
Post-training with preference optimization is critical
to align large language models with human preferred characteristics. From online alignment with  PPO~\citep{schulman2017ppo} to recent developments of offline alignment with DPO~\citep{rafailov2023direct,li2024kto,meng2024simpo}, LLMs improve by learning nuanced distinctions and separability between ranked responses. To align a learner LLM with human preferred characteristics such as helpfulness and safety, a pair of chosen and rejected responses generated by stronger teacher LLMs are ranked by either human-annotators~\citep{hh-rlhf2022Bai, wang-etal-2024-helpsteer, helpsteer2Wang2024} or LLM Judge~\citep{cui2024ultrafeedback}. However as the learner LLM's capabilities evolved, high-quality outputs for constructing strong preference data are hard to collect. \citep{geng2025delta} proposed delta learning which shows that preference optimization can effectively improve the learner LLM from the paired data created by weaker LLM as long as a meaningful quality gap can be constructed. While delta learning introduces a cost-effective approach for constructing high-quality preference data to improve instruct LLM's capabilities on general chat, the preference "delta" remains poorly understood in the context of reasoning. In this paper, we take a step further to systematically investigate delta learning on reasoning LLMs against verifiable tasks. 

\textbf{Preference Optimization for Reasoning Tasks. }
Recently there is a growing interest to extend preference optimization to enhance LLM's performance on objective reasoning domains like math~\citep{deepseek-math, pang2024iterative} and coding~\citep{zhang-li-2025-focused-dpo}. Unlike the general chat domains, reasoning tasks are often provided with golden answer to verify the correctness of model outputs. The preference pairs of reasoning tasks are constructed via contrasting a reasoning trace that reaches the correct final answer against the one that yields an incorrect answer. To handle the complexity of long reasoning chains, more recent research have focused on granular step-level optimization. \citep{lai2024stepdpo} applies step-wise preference optimization that focuses on fixing the initial logical deviations in long-chain reasoning, while~\citep{xu-etal-2025-full} densely annotates the contribution of each step to the final correct answer to learn the fine-grained level reasoning capabilities.
Inspired by the original delta learning work~\citep{geng2025delta}, we aim to study the effective delta to construct high-quality reasoning trace pairs for preference tuning beyond simply checking the outcome correctness in the reasoning domain. \citep{yao2025varying} initiate the effort to construct preference pairs between reasoning traces that are both verifiably wrong, but the chosen one contains less mistakes. This ``wrong-over-wrong'' preference pairs leads to gains on tasks of Knowledge Crosswords and biography generation. Our work distinguishes from this effort by systematically dissecting what constitutes a highly effective preference delta for reasoning. Instead of merely pairing correct and incorrect answers, we empirically investigate how the different properties of the generated traces such as cross-model family diversity, generator scale, and the absolute gap in reasoning trace quality drive downstream reasoning.



\textbf{Measuring Reasoning Trace Quality. }
Previous works on evaluating reasoning trace quality have studied the problem both at the level of the entire reasoning chain \citep{golovnevaroscoe, he2024socreval, prasad2023receval} and at the level of individual steps \citep{lightman2023let,zeng2024mrben,zeng2025mrgsmk,zheng2025processbench}.  Existing evaluation methods typically rely on critic models (LLM-as-a-judge) or reward models to score reasoning trace quality \citep{lee2025evaluating, he2024socreval, lightman2023let}. Recent works suggest that strong critic models can outperform reward models in reasoning quality evaluation \citep{zheng2025processbench, he2025can}. Beyond correctness, the proposed dimensions of assessing trace quality include informativeness \citep{prasad2023receval}, redundancy \citep{xia2025evaluating}, and coherence and relevance \citep{do2025defines}. \citet{lee2025evaluating} organizes these criteria into categories of factuality, validity (computational precision), coherence, and utility (signal-to-noise), which forms the basis of our reasoning quality evaluations.

\section{Conclusion}

We decomposed the quality gap in preference pairs for reasoning into two complementary notions: generator-level delta, the capability gap between the models that produce chosen and rejected responses, and sample-level delta, the quality difference within an individual pair. At the generator level, scaling the capability gap steadily improves out-of-domain generalization even as in-domain performance plateaus. At the sample level, we found that the learning signal resides not in outcome correctness but in the quality of the reasoning process---particularly step coherence---and that selecting for large sample-level deltas enables training on a fraction of the data with no loss in performance. Together, these results point to a practical recipe: maximize generator-level delta when constructing preference data, and exploit sample-level delta to select the most informative examples from it.


\section*{Ethics Statement}
This work investigates the properties of preference pairs that drive reasoning improvements in language models, using only a publicly available dataset (i.e., OpenR1-Math-220k) and open-weight models for response synthesis. No human subjects were involved in data collection or annotation. All preference pairs were constructed from model-generated reasoning traces, and quality evaluations were performed by an LLM judge (GPT-OSS-120b) rather than human annotators. We acknowledge that our findings on early gains from limited data could lower the computational cost of alignment, which we view as a positive societal outcome, but we also recognize that improvements in reasoning capabilities could in principle be applied to both beneficial and harmful ends. We do not foresee direct negative societal impacts from the methodological contributions of this paper, as our work focuses on understanding existing training paradigms rather than introducing new model capabilities. In accordance with the COLM policy on LLM usage, we disclose that LLMs were used for data generation, evaluation (as reasoning trace quality judges), and minor assistance in paper writing.

\section*{Acknowledgment}
We thank Berkcan Kapusuzoglu for providing an early version of evaluation scripts.

\bibliography{colm2026_conference}
\bibliographystyle{colm2026_conference}

\appendix
\section{Additional Experimental Setup Details}
\subsection{Trace Quality Judge Setup Details}
\label{app:trace_judge_setup}
This section details the prompt architecture and scoring rubric used for GPT-OSS-120b ratings of reasoning trace quality.

\subsubsection{Comprehensive Evaluation Rubric}

The following table defines the dimensions across the two primary categories: \textbf{Logic} and \textbf{Brevity}.

\begin{table}[h!]
    \centering
    \small
    \begin{tabularx}{\textwidth}{l p{2.2cm} X X}
        \toprule
        \textbf{Cat.} & \textbf{Dimension} & \textbf{Success Criteria} & \textbf{Deduction Triggers} \\ \midrule
        \textbf{Log.} & Factuality & Correctly identifies constants, variables, and constraints. No hallucinations. & Hallucinating data, ignoring explicit constraints. \\ \addlinespace
        & Strategy & Approach is appropriate for problem type; follows coherent sequence. & Inefficient/incorrect theorem, circular logic. \\ \addlinespace
        & Step Coherence & Valid mathematical transitions; each step follows logically from the previous. & Unjustified leaps, 'magic' numbers, invalid transformations. \\ \addlinespace
        & Computational Precision & Zero errors in arithmetic, simplification, or symbolic expansion. & Arithmetic errors, sign errors, typos. \\ \midrule
        \textbf{Brev.} & Signal-to-Noise & High density of informative steps; minimal meta-commentary or redundancy. & Excessive meta-commentary, irrelevant digressions, circular work. \\ \bottomrule
    \end{tabularx}
\end{table}

\subsubsection{System Prompt and Instructions}

\begin{PromptBox}{GPT-OSS-120b Trace Quality Rater}
    \textbf{Persona:} You are an expert Mathematical Logic Auditor. Your task is to evaluate the Logic and Brevity of a reasoning trace for a given math problem. Your sole purpose is to critique the provided Reasoning Trace based on the structured rubric. \\
    
    \textbf{Constraint:} DO NOT provide your own solution to the Problem Statement. Even if the reasoning trace is long or incomplete, simply evaluate what is there using the rubric.
    
    \vspace{0.5em}
    \textbf{Scoring Logic (1--5 Scale):}
    \begin{itemize}[leftmargin=1.5em, nosep]
        \item \textbf{5 (Excellent):} Meets all Success Criteria; no Deduction Triggers present.
        \item \textbf{3 (Mediocre):} Mostly meets criteria but contains a significant Deduction Trigger.
        \item \textbf{1 (Fail):} Entirely fails the Dimension (e.g., incorrect method, filler).
    \end{itemize}
    \textit{Requirement: Provide a one-sentence justification and list all present Deduction Triggers.}
\end{PromptBox}

\begin{CodeBox}{Output JSON Schema Format}
\{ \\
\quad "\{category\_name\}": \{ \\
\quad \quad "\{dimension\_name\}": \{ \\
\quad \quad \quad "score": (integer 1-5), \\
\quad \quad \quad "justification": "string", \\
\quad \quad \quad "deductions\_found": ["string from triggers"] \\
\quad \quad \} \\
\quad \} \\
\}
\end{CodeBox}

We use guided decoding to enforce that GPT-OSS-120b generates ratings following the specified JSON format for accurate parsing.

\section{Data Overlap Across Dimensions}
\label{app:dim_overlap}
In Table \ref{tab:scaling_quality} in the main paper, we observed that the average delta between chosen samples from s1-32B and rejected samples from s1-3B is large in all dimensions.  This raises a natural question: to what extent do the datasets formed by selecting the top 1000, 2000, and 5000 samples according to deltas in each dimension overlap with one another?

\begin{figure}[ht]
    \centering
    \includegraphics[width=\linewidth]{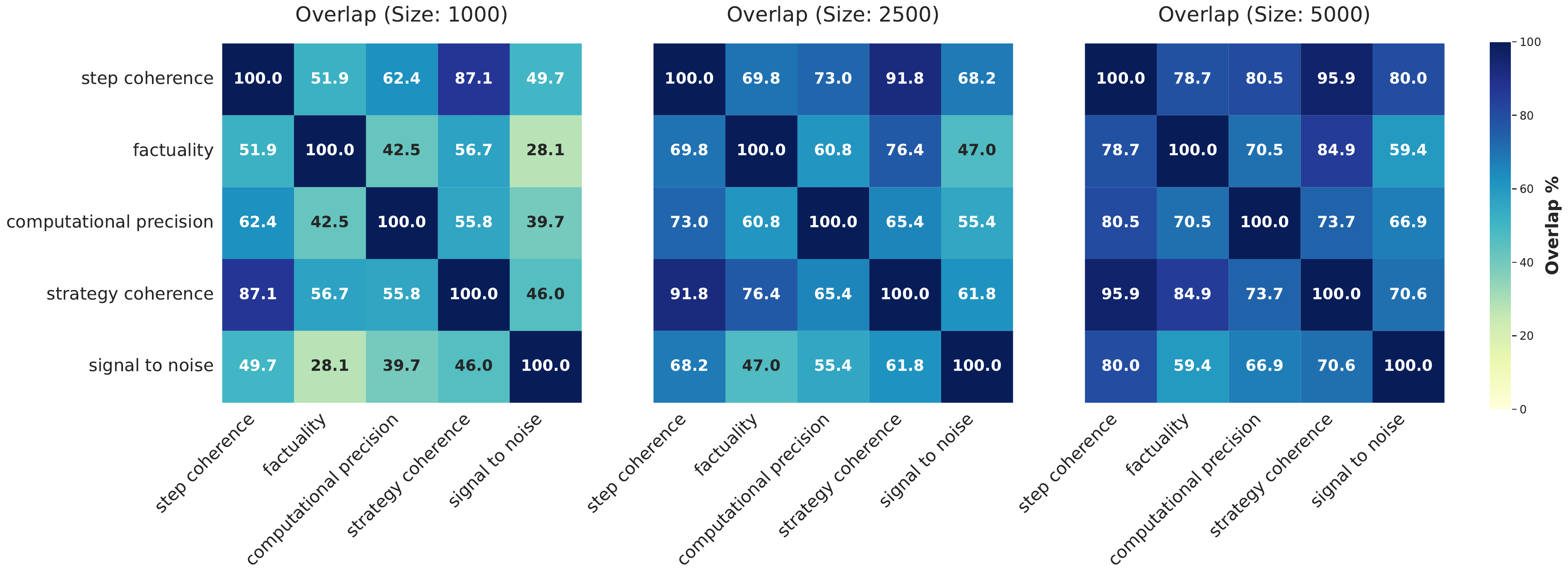}
    \caption{Overlap between data in datasets formed by top 1000, 2500, and 5000 deltas in each dimension.}
    \label{fig:data_overlap_distribution}
\end{figure}

Figure \ref{fig:data_overlap_distribution} visualizes the percentage overlap between pairs of datasets constructed from the top-ranked examples for each dimension.  Overall, we find degree of overlap increases with dataset size.  We observe that some dimensions contain substantial overlap, suggesting that these qualities are correlated for s1-32B and s1-3B response pairs.  For example, strategy coherence and step coherence have a large overlap even at the smallest size set (87.1\% at 1k examples).  In contrast, the overlap between factuality and signal-to-noise at 1k samples is much lower (28.1\%) which suggests a lower degree of correlation between these dimensions.

\section{Off-the-shelf Model Performance across Benchmarks}
\label{app:benchmark-comparison}
Table~\ref{tab:benchmark-comparison} shows the performance for the off-the-shelf models we used across mathematical reasoning, general knowledge, and code generation benchmarks. For AMC23, AIME24, AIME25, we calculate the average accuracy across 16 reasoning traces.

\begin{table}[]
\centering
\footnotesize
\setlength{\tabcolsep}{2.5pt}
\renewcommand{\arraystretch}{1.25}
\resizebox{\columnwidth}{!}{%
\begin{tabular}{lccccccccccc}
\hline
Model
& MATH
& Minerva
& GSM
& Olympiad
& AMC23
& AIME24
& AIME25
& TheoremQA
& GPQA
& MMLU-Pro
& LiveCode \\
\hline
Nemotro-4B  & 93.3 & 55.1 & 92.0 & 60.3 & 91.4 & 56.5 & 44.6 & 55.9 & 31.0 & 51.7  & 11.2 \\
Nemotro-8B  & 91.2 & 52.9 & 87.6 & 61.0 & 93.3 & 61.0 & 50.8 & 55.6 & 46.6 & 57.7 & 12.1 \\
Nemotro-49B & 94.2 & 57.4 & 95.0 & 63.9 & 94.2 & 60.4 & 52.9 & 63.2 & 62.3 & 77.0  & 27.7    \\
\hline
S1 3B       & 61.0 & 24.6 & 86.4 & 24.7 & 35.9 & 4.8  & 5.6  & 36.2 & 16.1    & --    & 3.6  \\
S1 7B       & 80.0 & 45.2 & 91.5 & 44.1 & 59.8 & 18.5 & 20.6 & 51.4 & 28.3    & --    & 4.8  \\
S1 14B      & 88.4 & 52.9 & 95.0 & 53.5 & 77.2 & 34.8 & 29.8 & 58.6 & 48.9    & --    & 7.9    \\
S1 32B      & 92.6 & 61.4 & 94.8 & 62.7 & 93.4 & 55.0 & 47.3 & 65.9 & 50.9    & --    & --    \\
\hline
\end{tabular}
}
\caption{Benchmark results for off-the-shelf models.}
\label{tab:benchmark-comparison}
\end{table}

\section{Training Details}
\label{app:training_details}

For the alignment phase, we utilize a maximum context length of 8192 tokens across all training configurations. Models are trained until the objective loss reaches convergence; however, the final model selection is determined by identifying the best-performing checkpoint immediately prior to convergence to mitigate potential over-optimization.

The specific hyperparameter configurations for Direct Preference Optimization (DPO) and Kahneman-Tversky Optimization (KTO) are detailed below:

\begin{itemize}
    \item \textbf{Direct Preference Optimization (DPO):} We employ a learning rate of $5 \times 10^{-8}$ and a regularization parameter $\beta = 0.2$.
    \item \textbf{Kahneman-Tversky Optimization (KTO):} We utilize a learning rate of $5 \times 10^{-8}$ with $\beta = 0.5$. The weights for desirable and undesirable outcomes are both set to $1.0$.
\end{itemize}

\end{document}